%% file: main.tex
\documentclass[11pt,a4paper]{article}
\usepackage{emnlp2018}
\usepackage{times}
\usepackage{latexsym}
\usepackage[utf8]{inputenc}
\usepackage{graphicx}
\usepackage{tikz}
\usepackage{multirow}
\usepackage{amsmath}
\usepackage{amssymb}
\usepackage{todonotes}
\usepackage{standalone}
\usepackage{booktabs}

\usepackage{url}

\usetikzlibrary{shapes}
\aclfinalcopy 

\DeclareMathOperator{\attn}{\mathcal{A}}

\DeclareMathOperator*{\softmax}{softmax}

\title{CUNI System for the WMT18 Multimodal Translation Task}

\author{Jindřich Helcl \and Jindřich Libovický \and Dušan Variš \\
  Charles University, Faculty of Mathematics and Physics \\
  Institute of Formal and Applied Linguistics \\
  Malostransk\' e n\' am\v est\' i 25, 118 00 Prague, Czech Republic \\
  {\tt \{helcl, libovicky, varis\}@ufal.mff.cuni.cz}}

\date{}

\begin{document}
\maketitle
\begin{abstract}
We present our submission to the WMT18 Multimodal Translation Task.
The main feature of our submission is applying a self-attentive network instead of 
a recurrent neural network.
%
%
We evaluate two methods of incorporating the visual features in the model: first, we
include the image representation as another input to the network; second, we train the
model to predict the visual features and use it as an auxiliary objective.
For our submission, we acquired both textual and multimodal additional data.
Both of the proposed methods yield significant improvements over recurrent networks and self-attentive textual baselines.
%
%
%
%
\end{abstract}

\section{Introduction}



Multimodal Machine Translation (MMT) is one of the tasks that seek ways of capturing
the relation of texts in different languages given a shared ``grounding'' information
in a different (e.g. visual) modality.

The goal of the MMT shared task is to generate an image description (caption)
in the target language using a caption in the source language and the image itself.
The main motivation for this task is the development of models that can exploit
the visual information for meaning disambiguation and thus model the denotation of words.

During the last years, MMT was addressed as a subtask of neural machine translation
(NMT). It was thoroughly studied within the framework of recurrent neural networks
(RNNs) \citep{specia2016shared,elliott2017findings}. Recently, the architectures
based on self-attention such as the Transformer \citep{vaswani2017attention} became
state-of-the-art in NMT.

In this work, we present our submission based on the Transformer model.
We propose two ways of extending the model. First, we tweak the architecture
such that it is able to process both mo\-da\-li\-ties in a multi-source learning scenario.
Second, we leave the model architecture intact, but add another training objective
and train the textual encoder to be able to predict the visual features of the image
described by the text. This training component has been introduced in RNNs by
\citet{elliot2017imagination} and is called the ``imagination''.

We find that with self-attentive networks, we are able to improve over a strong
textual baseline by including the visual information in the model. This has been
proven challenging in the previous RNN-based submissions, where there was only a minor
difference in performance between textual and multimodal models \citep{helcl2017cuni,caglayan2017lium}.

This paper is organized as follows. Section~\ref{sec:related} 
summarizes the previous submissions and related work. In Section~\ref{sec:architecture},
we describe the proposed methods. The details of the datasets used for the training
are given in Section~\ref{sec:data}. Section~\ref{sec:experiments} describes the 
conducted experiments. We discuss the results in Section~\ref{sec:results} and 
conclude in Section~\ref{sec:conclusions}.

\section{Related Work}
\label{sec:related}


Currently, most of the work has been done within the framework of sequence-to-sequence 
learning. Although some of the proposed approaches use
explicit image analysis \citep{shah2016shef,huang2016attention}, most methods use image
representation obtained using image classification networks pre-trained on ImageNet \citep{deng2009imagenet}, usually VGG19~\cite{simonyan2014vgg} or ResNet~\citep{he2016deep}.


In the simplest case, the image can be represented as a single vector from the penultimate
layer of the image classification network. This vector can be then plugged in at 
various places of the sequence-to-sequence architecture
\citep{libovicky2016cuni,calixto2017incorporating}.

Several methods compute visual context information as a weighted sum over the image
spatial representation using the attention mechanism 
\citep{bahdanau2015neural,xu2015show} and combine it with the context vector from
the textual encoder in doubly-attentive decoders. 
\citet{caglayan2016multimodality} use the visual context vector in a gating mechanism 
applied to the textual context vector. \citet{caglayan2017lium} concatenate the context 
vectors from both modalities. \citet{libovicky2017attention} proposed advanced
strategies for computing a joint attention distribution over the text and image.
We follow this approach in our first proposed method described in 
Section~\ref{ssec:doubly}.

The visual information can also be used as an auxiliary objective in a multi-task 
learning setup. \citet{elliot2017imagination} propose an imagination component that 
predicts the visual features of an image from the textual encoder representation, 
effectively regularizing the encoder part of the network. The imagination component
is trained using a maximum margin objective. We reuse this approach in our method 
described in Section~\ref{ssec:imagination}.

\section{Architecture}
\label{sec:architecture}



We examine two methods of exploiting the visual information in the Transformer
architecture. First, we add another encoder-decoder attention layer to the decoder
which operates over the image features directly.
Second, we train the network with an auxiliary objective using the imagination 
component as proposed by \citet{elliot2017imagination}.

\subsection{Doubly Attentive Transformer}
\label{ssec:doubly}



The Transformer network follows the encoder-decoder scheme. Both parts consist
of a number of layers. Each encoder layer first attends to the previous layer using
self-attention, and then applies a single-hidden-layer feed-forward network to the
outputs. All layers are interconnected with residual connections and their outputs
are normalized by layer normalization \citep{ba2016layernorm}. A decoder layer differs
from an encoder layer in two aspects. First, as the decoder operates 
autoregressively, the self-attention has to be masked to prevent the decoder
to attend to the ``future'' states. Second, there is an additional attention sub-layer
applied after self-attention which attends to the final states of the encoder (called \emph{encoder-decoder}, or \emph{cross} attention). 

%

The key feature of the Transformer model is the use of attention mechanism instead 
of recurrence relation in RNNs.
The attention can be conceptualized as a soft-lookup function that operates on an 
associative array. For a given set of queries $Q$, the attention uses a similarity
function to compare each query with a set of keys $K$. The resulting similarities
are normalized and used as weights to compute a context vector which is a weighted
sum over a set of values $V$ associated with the keys. 
In self-attention, all the queries, keys and values correspond to the set of states
of the previous layer. In the following cross-attention sub-layer, the set of
resulting context vectors from the self-attention sub-layer is used as queries,
and keys and values are the states of the final layer of the encoder.

The Transformer uses scaled dot-product as a similarity metric for both
self-attention and cross-attention. For a query matrix $Q$, key matrix $K$ and value
matrix $V$, and the model dimension $d$, we have:
\begin{gather}
 \attn(Q, K, V) = \softmax \left( \dfrac{QK^\top}{\sqrt{d}}\right)V.
\end{gather}

The attention is used in a multi-head setup. This means that we first linearly project 
the queries, keys, and values into a number of smaller matrices and then apply the
attention function $\mathcal{A}$ independently on these projections.
The set of resulting context vectors $C$ is computed as a sum of the outputs of each
attention head, linearly projected to the original dimension:
\begin{gather}
C = \sum_{i=1}^h \attn(QW^Q_i, KW^K_i, VW^V_i) W^O_i   
%
%
%
\end{gather}
where $(W^O_i)^{\top}$, $W^Q_i$, $W^K_i$, and $W^V_i \in \mathbb{R}^{d \times d_h}$ are
trainable parameters, $d$ is the dimension of the model, $h$ is the number of heads,  
and $d_h$ is a dimension of a single head. Note that despite $K$ and $V$ being identical
matrices, the projections are trained independently.

\begin{figure}
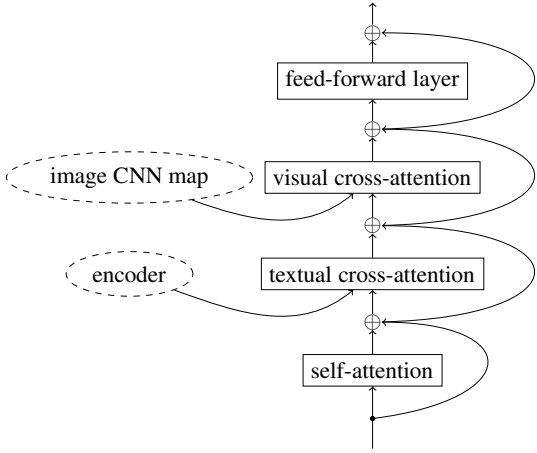

    \centering
    \includestandalone[width=\linewidth]{./img/serial}
    \caption{One layer of the doubly-attentive Transformer decoder with 4
    sub-layers connected with residual connections.}
    \label{fig:serial}
\end{figure}

In this method, we introduce the visual information to the model as another encoder
via an additional cross-attention sub-layer. The keys and values of this
cross-attention correspond to the vectors in the last convolutional layer of a
pre-trained image processing network applied on the input image. This sub-layer is
inserted between the textual cross-attention and the feed-forward network, as 
illustrated in Figure~\ref{fig:serial}. The set of the context vectors from the
textual cross-attention is used as queries, and the context vectors of the visual
cross-attention are used as inputs to the feed-forward sub-layer.
Similarly to the other sub-layers, the input is linked to the output by a residual 
connection. Equation~\ref{eq:cimg} shows the computation of the visual context vectors
given trainable matrices $Z_i^Q$, $Z_i^K$, $Z_i^V$, and $Z_i^O$ for $i = 1, \ldots, h$;
the set of textual context vectors is denoted by $C_\mathit{txt}$ and the extracted 
set of image features as $F$:
\begin{gather}
    C_\mathit{img} = \sum_{i=1}^h \attn(C_\mathit{txt}Z^Q_i, FZ^K_i, FZ^V_i) Z^O_i.
    \label{eq:cimg}
\end{gather}



\subsection{Imagination}
\label{ssec:imagination}



We use the imagination component of \citet{elliot2017imagination} originally
proposed for training multimodal translation models using RNNs.
We adapt it in a straightforward way in our Transformer-based models.

The imagination component serves effectively as a regularizer to the encoder,
making it consider the visual meaning together with the words in the source sentence.
This is achieved by training the model to predict the image representations that
correspond to those computed by a pre-trained image classification network.
Given a set of encoder states $h_j$, the model computes the predicted image 
representation as follows:
\begin{gather}
    \hat{y}_\mathit{img} = W^R_2 \, \max(0, W^R_1 \, {\textstyle \sum_j} h_j)
    \label{eq:regressor}
\end{gather}
where $W_1^R \in \mathbb{R}^{r \times d} $ and $W_2^R \in \mathbb{R}^{n \times r}$ 
are trainable parameter matrices, $d$ is the Transformer model dimension, $r$ is a 
hidden layer dimension of the imagination component, and $n$ is the dimension of 
the image feature vector. Note that Equation~\ref{eq:regressor} corresponds to
a single-hidden-layer feed-forward network with a ReLU activation function 
applied on the sum of the encoder states.

We train the visual feature predictor using an auxiliary objective. Since the encoder part
of the model is shared, additional weight updates are propagated to the encoder during the
model optimization w.r.t. this additional loss.
For the generated image representation $\hat{y}$ and the reference representation $y$,
the error is estimated as margin-based loss with margin parameter
$\alpha$:
\begin{equation}
    \mathcal{L}_\mathit{imag} = \max \left( 0, \alpha + d(\hat{y}, y) - d(\hat{y}, y_c) \right)
    \label{eq:hingeloss}
\end{equation}
where $y_c$ is a contrastive example randomly drawn from the training batch
and $d$ is a distance function between the representation vectors, in our case 
the cosine distance.

Unlike \citet{elliot2017imagination}, we sum both translation and imagination losses
within the training batches rather than alternating between training of each component separately.

\begin{table}
\begin{center}
\begin{tabular}{lcccc}
\toprule
& en & de & fr & cs\\ 
\midrule
Training & \multicolumn{4}{c}{29,000 sentences} \\ 
\midrule
Tokens &  378k & 361k & 410k & 297k \\
Average length &  13.0 & 12.4 & 14.1 & 10.2 \\
\# tokens range & \small 4--40 & \small 2--44 & \small 4--55 & \small 2--39 \\ 
\midrule
Validation & \multicolumn{4}{c}{1,014 sentences} \\ 
\midrule
Tokens &  13k & 13k & 14k & 10k \\
Average length & 13.1 & 12.7 & 14.2 & 10.2  \\
\# tokens range & \small 4--30 & \small 3--33 & \small 5--36 & \small 4--27 \\
\midrule 
OOV rate & \small 1.28\% & \small 3.09\% & \small 1.20\% & \small 3.95\% \\
\bottomrule
\end{tabular}
\end{center}

\caption{Multi30k statistics on training and validation data -- total number of
tokens, average number of tokens per sentence, and lengths of the shortest
and the longest sentence.}
\label{tab:data}
\end{table}

\section{Data}
\label{sec:data}



The participants were provided with the Multi30k dataset \citep{elliott2016multi30k}, an extension of the Flickr30k dataset \citep{plummer2017flickr30k} which 
contains 29,000 train images, 1,014 validation images and 1,000 test images. The
images are accompanied with six captions which were independently obtained through crowd-sourcing.
In Multi30k, each image is accompanied also with German, French, and Czech translations
of a single English caption. Table~\ref{tab:data} shows statistics of the captions
contained in the Multi30k dataset.

Since the Multi30k dataset is relatively small, we acquired additional data, similarly to our
last year submission \citep{helcl2017cuni}. The overview of the dataset structure is
given in Table~\ref{tab:additional_data}. 

First, for German only, we prepared synthetic data out of the WMT16 MMT Task~2 training dataset
using back-translation to English \citep{sennrich2016backtranslation}.
This data consists of five additional German descriptions of each image.
Along with the data for Task~1 which is the same as the training data this year, the 
back-translated part of the dataset contains 174k sentences. 

Second, for Czech and German, we selected pseudo in-domain data by filtering the available
general domain corpora. For both languages, we trained a character-level RNN language model
on the corresponding language parts of the Multi30k training data. We use a single layer
bidirectional LSTM \cite{hochreiter1997lstm} network with 512 hidden units and character
embeddings with dimension of 128. For Czech, we compute perplexities of the Czech sentences
in the CzEng corpus \citep{bojar2016czeng}. We selected 15k low-perplexity sentence pairs 
out of 64M sentence pairs in total by setting the perplexity threshold to 2.5. For German,
we used the additional data from the last year \citep{helcl2017cuni}, which was selected out of 
several parallel  corpora (EU Bookshop~\citep{tiedemann2014billions}, News Commentary~\citep{tiedemann2012parallel} and CommonCrawl~\citep{smith2013dirt}).

Third, also for Czech and German, we applied the same criterion on monolingual corpora
and used back-translation to create synthetic parallel data. For Czech, we took 
333M sentences of CommonCrawl and 66M sentences of News Crawl (which is used in the WMT 
News Translation Task; \citealp{bojar2016findings}) and extracted 18k 
and 11k sentences from these datasets respectively.

Finally, we use the whole EU Bookshop as an additional out-of-domain parallel data.
Since the size of this dataset is large relative to the sizes of the other parts, we oversample
the rest of the data to balance the in-domain and out-of-domain portions of the training dataset.
The oversampling factors are shown in Table~\ref{tab:additional_data}.

For the unconstrained training of the imagination component, we used the MSCOCO 
\citep{tsung2014mscoco} dataset which consists of 414k images along with English captions.

\begin{table}
\begin{tabular}{lccc}
\toprule
                   &   de  &   fr  &   cs   \\
\midrule 
Multi30k               & \multicolumn{3}{c}{29k}\\
\small -- oversampling factor & \small 273$\times$ & \small 366$\times$ & \small 9$\times$ \\

\midrule 
Task 2 BT              &  145k &   --- &   ---  \\
in-domain parallel     &    3k &   --- &   15k  \\
in-domain BT           &   30k &   --- &   29k  \\
\small -- oversampling factor & \small 39$\times$ & --- & \small 7$\times$  \\
\midrule
EU Bookshop            &  9.3M & 10.6M &  445k  \\ 
\midrule
COCO (English only)    & \multicolumn{3}{c}{414k} \\
\bottomrule
\end{tabular}

\caption{Overview of the data used for training our models with oversampling factors. The 
EU Bookshop data was not oversampled. BT stands for back-translation.}
\label{tab:additional_data}
\end{table}

\begin{table*}
    \centering
    \begin{tabular}{llcccccc}
    \toprule
     & & \multicolumn{2}{c}{en-cs} & \multicolumn{2}{c}{en-fr} & \multicolumn{2}{c}{en-de} \\ 
     & & single & averaged & single & averaged & single & averaged \\
     \midrule
     \multicolumn{2}{l}{\citet{caglayan2017lium}} & 
        \multicolumn{2}{c}{N/A} & 
        54.7/71.3 & 56.7/73.0 &
        37.8/57.7 & 41.0/{\bf 60.5}
        \\ \midrule
         \multirow{3}{*}{\rotatebox{90}{Cons.}} 
         & Textual  &
            29.6/28.9 & 30.9/29.5 &   
            59.2/73.7 & 59.7/74.4 &   
            38.1/56.2 & 38.3/56.0     
            \\
         & Imagniation &
            29.8/29.4 & 30.5/29.6 &   
            59.4/74.2 & 59.7/74.4 &   
            38.8/56.4 & {\bf}39.2/56.8     
            \\
         & Multimodal &
            30.5/29.7 & {\bf}31.0/29.9 &   
            60.6/75.0 & {\bf}60.8/75.1 &   
            38.4/53.1 & 38.7/{\bf}57.2     
           \\ \midrule
         \multirow{2}{*}{\rotatebox{90}{Unc.}}
         & Textual &
            31.2/30.1 & 32.3/30.7 &   
            62.0/76.7 & 62.5/76.7 &   
            39.6/58.7 & 40.4/59.0     
            \\
         & Imagination &
            \bf 36.3/32.8 & 35.9/32.7 &   
            \bf 62.8/77.0 & \bf 62.8/77.0 &   
            {\bf 42.7}/59.1 & 42.6/59.4     
            \\
    \bottomrule
    \end{tabular}
    \caption{Results on the 2016 test set in terms of BLEU score and METEOR score. We compare
             our results with the last year's best system \citep{caglayan2017lium}
             which used model ensembling instead of weight averaging.}
    \label{tab:results}
\end{table*}

\section{Experiments}
\label{sec:experiments}



In this year's round, two variants of the MMT tasks were announced. As in the
previous years, the goal of Task~1 is to translate an English caption into
the target language given the image. The target languages are German, French and 
Czech. In Task~1a, the model receives the image and its captions in English, 
German, and French and is trained to produce the Czech translation.
In our submission, we focus only on Task~1.

In our submission, we experiment with three distinct architectures. First, in 
\emph{textual} architectures, we leave out the images from the training altogether.
We use this as a strong baseline for the multimodal experiments. Second, \emph{multimodal} 
experiments use the doubly attentive Transformer decoder described in Section~\ref{ssec:doubly}.
Third, the experiments referred to as \emph{imagination} employ the imagination component
as described in Section~\ref{ssec:imagination}.

We train the models in constrained and unconstrained setups. In the 
constrained setup, only the Multi30k dataset is used for training. In the 
unconstrained setup, we train the model using the additional data described in 
Section~\ref{sec:data}. We run the multimodal experiments only in the constrained setup. 

In the unconstrained variant of the imagination experiments, the dataset consists of examples
that can miss either the textual target values (MSCOCO extension), or the image 
(additional parallel data). In these cases, we train only the decoding component
with specified target value (i.e. imagination component on visual features, or the 
Transformer decoder on the textual data). As said in Section~\ref{ssec:imagination},
we train both components by summing the losses when both the image and the target sentence are
available in a training example.

In all experiments, we use the Transformer network with 6 layers with model
dimension of 512 and feed-forward hidden layer dimension of 4096 units. The embedding
matrix is shared between the encoder and decoder and its transposition is reused as the
output projection matrix \citep{press2017tieembeddings}. For each language pair, we use a
vocabulary of approximately 15k wordpieces \citep{wu2016google}. We extract the vocabulary
and train the model on lower-cased text without any further pre-processing steps applied.
We tokenize the text using the algorithm bundled with the tensor2tensor library
\citep{tensor2tensor}. The tokenization algorithm splits the sentence to groups of alphanumeric and non-alphanumeric groups, throwing away single spaces that occur inside the sentence.
We conduct the experiments using the Neural Monkey toolkit \citep{NeuralMonkey:2017}.\footnote{\url{https://github.com/ufal/neuralmonkey}}

For image pre-processing, we use ResNet-50 \citep{he2016deep} with identity mappings 
\citep{he2016identity}. In the doubly-attentive model, we use the outputs of the
last convolutional layer before applying the activation function with
dimensionality of $8 \times 8 \times 2048$. We apply a trainable linear projection
to the maps into 512 dimensions to fit the Transformer model dimension. In the imagination
experiments, we use average-pooled maps with 2048 dimensions. Following 
\citet{elliot2017imagination}, we set the margin parameter $\alpha$ from 
Equation~\ref{eq:hingeloss} to 0.1.

For each model, we keep 10 sets of parameters that achieve the best BLEU scores
\citep{papineni2002bleu} on the validation set. We experiment with weight averaging and model ensembling. However, these methods performed similarly and we thus report only the results
of the weight averaging, which is computationally less demanding.

In all experiments, we use the Adam optimizer \citep{kingma2015adam} with initial learning
rate 0.2, and Noam learning rate decay scheme \citep{vaswani2017attention} with $\beta_1$ = 0.9, 
$\beta_2 = 0.98$ and $\epsilon = 10^{-9}$ and 4,000 warm-up steps.

\section{Results}
\label{sec:results}



We report the quantitative results of measured on the Multi30k 2016 test set in
Table~\ref{tab:results}.

The Transformer architecture achieves generally comparable or better results than the
RNN-based architecture. Adding the visual information has a significant positive effect
on the system performance, both when explicitly provided as a model input and when used
as an auxiliary objective. In the constrained setup which used only the data from 
the Multi30k dataset, the doubly-attentive decoder performed best.

The biggest gain in performance was achieved by training on the additional parallel data.
The imagination architecture outperforms the purely textual models.

As the performance of single models increases, the positive effect of weight averaging
diminishes. The effect of checkpoint averaging is smaller than the results
reported by \citet{caglayan2017lium} who use ensembles of multiple models
trained with a different initialization -- we use only checkpoints from a single training run.

During the qualitative analysis, we noticed that mostly for Czech target language, 
the systems are often incapable of capturing morphology. In order to quantify this, we also
measured the BLEU scores using the lemmatized system outputs and references. 
The difference was around 4 BLEU points for Czech, less than 3 BLEU points for French,
and around 2 BLEU points for German. These differences were consistent among different 
types of models.

We hypothesize that in the imagination experiments, the visual information is used
to learn a better representation of the textual input, which eventually leads to
improvements in the translation quality. In the multimodal experiments, the improvements
can come from the refining of the textual representation rather than from explicitly using
the image as an input. 

In order to determine whether the visual information is used also at the inference time, 
we performed an adversarial evaluation by providing the trained multimodal model with
randomly selected ``fake'' images.
In French and Czech, BLEU scores dropped by more than 1 BLEU point. This suggests that the
multimodal models utilize the visual information at the inference time as well.
The German models seem to be virtually unaffected. We hypothesize this might be due to
a different methodology of acquiring the training data for German and the other two
target languages \citep{elliott2016multi30k}.

\section{Conclusions}
\label{sec:conclusions}



In our submission for the WMT18 Multimodal Translation Task, we experimented with the 
Transformer architecture for MMT. The experiments show that the Transformer 
architecture outperforms the RNN-based models.

Experiments with a doubly-attentive decoder showed that explicit incorporation of visual
information improves the model performance. The adversarial evaluation confirms that the
models also take into account the visual information.

The best translation quality was achieved by extending the training data 
by additional image captioning data and parallel textual data.  
It this unconstrained setup, the best scoring model employs the imagination
component that was previously introduced in RNN-based sequence-to-sequence models.

\section*{Acknowledgements}

This research received support from
the Czech Science Foundation grant no. P103/12/G084,
and the grants No. 976518 and 1140218 of the Grant Agency of the Charles University.
This research was partially supported by SVV project number 260~453.

\bibliography{mmt}
\bibliographystyle{acl_natbib_nourl}
\end{document}

%% file: img/serial.tex
\def\verticalstep{0.8}

\begin{tikzpicture}[]

\draw[fill,outer sep=0,inner sep=0]  (0,0) circle (1pt) node (before_self) {};
\draw (0,-0.5) -- (before_self);

\draw (0,1 * \verticalstep) node[rectangle,draw] (self_att) {self-attention};
\draw[inner sep=0]  (0,2 * \verticalstep) node (after_self) {$\oplus$};

\draw[->] (before_self) -- (self_att);
\draw[->] (before_self) .. controls ++(2.5,0.4 * \verticalstep) and ++(2.5,0) .. (after_self);
\draw[->] (self_att) -- (after_self);

\draw (0,3 * \verticalstep) node[rectangle,draw] (cross_att_1) {textual cross-attention};
\draw (-4,3 * \verticalstep) node[shape=ellipse,draw,style=dashed] (encoder_1) {encoder};
\draw[inner sep=0]  (0,4 * \verticalstep) node (after_cross_1) {$\oplus$};

\draw[->] (encoder_1) to[out=340,in=220] (cross_att_1);
\draw[->] (after_self) -- (cross_att_1);
\draw[->] (cross_att_1) -- (after_cross_1);
\draw[->] (after_self) .. controls ++(3.5,0.1 * \verticalstep) and ++(3.5,0) .. (after_cross_1);

\draw (0,5  * \verticalstep) node[rectangle,draw] (cross_att_2) {visual cross-attention};
\draw (-4,5  * \verticalstep) node[shape=ellipse,draw,style=dashed] (encoder_2) {image  CNN map};
\draw[inner sep=0]  (0,6  * \verticalstep) node (after_cross_2) {$\oplus$};

\draw[->] (encoder_2) to[out=340,in=220] (cross_att_2);
\draw[->] (after_cross_1)-- (cross_att_2);
\draw[->] (cross_att_2) -- (after_cross_2);
\draw[->] (after_cross_1) .. controls ++(3.5,0.1 * \verticalstep) and ++(3.5,0) .. (after_cross_2);

\draw (0,7  * \verticalstep) node[rectangle,draw] (ff) {feed-forward layer};
\draw[inner sep=0]  (0,8  * \verticalstep) node (after_ff) {$\oplus$};

\draw[->] (after_cross_2) -- (ff);
\draw[->] (after_cross_2) .. controls ++(3.5,0.1 * \verticalstep) and ++(3.5,0) .. (after_ff);
\draw[->] (ff) -- (after_ff);

\draw[->] (after_ff) -- ++(0,0.5);

\end{tikzpicture}